\documentstyle[11pt,aclap]{article}

\newcommand{\comment}[1]{}
\newcommand{\done}[1]{}

\newcommand{\ssp}{\setlength{\baselineskip}{13pt}}

\newcommand{\SF}{SF}
\newcommand{\SFs}{SFs}
\newcommand{\SD}{SD}

\newcounter{examplectr}
\newcounter{subexamplectr}

\newenvironment{ex}%
   {\refstepcounter{examplectr}
     \setcounter{subexamplectr}{0}
     \begin{list}
       {(\arabic{examplectr})}%
       {\setlength{\topsep}{0in}
        \setlength{\leftmargin}{0.75in}
        \setlength{\labelsep}{0.15in}}
       \item \ssp \begin{minipage}[t]{4.75in}
       }%
   {\end{minipage}
    \end{list}}

\newcommand{\bex}{\begin{ex}}
\newcommand{\eex}{\end{ex}}

\newcommand{\bit}{\begin{itemize}}

\newcommand{\eit}{\end{itemize}}

\newcommand{\attr}[1]{{\small \tt #1}}

\newcommand{\tind}[1]{\fbox{{\tiny #1}}\, }

\newcommand{\avm}[1]{\mbox{\begin{math}
			   \setlength{\arraycolsep}{1mm}
                           \renewcommand{\arraystretch}{1.1}      
                           \hspace*{-0.35em}
			   \left[
                           \begin{array}{@{}l@{~}l@{}} 		
                             \\[-0.16in] #1 \\[-0.16in]           
                           \end{array}
                           \right] \hspace*{-0.05em}
                           \end{math}}}

\newcommand{\tavm}[2]{\mbox{{\begin{tabular}{@{}l@{}}
                                        $\mbox{\it #1 }^{\avm{#2}}$
                                 \end{tabular}}}}

\newcommand{\etavm}[1]{\raisebox{-0.9ex}{$\mbox{\it #1}^{\mbox{\rm{[\ \ ]}}}$}}

\newcommand{\avl}[2]{\mbox{\attr{#1}} & \it{#2}  \\ }

\newcommand{\eavm}{\ \avm{}}

\newcommand{\nel}[1]{\mbox{$\left\langle #1 \right\rangle$}}

\newcommand{\eyl}[0]{\mbox{$\left\langle ~ \right\rangle$}}

\newcommand {\treelabel}[1]{\tt #1}

\newcommand{\Sequence}[1]{<#1>}
\newcommand{\AVM}[1]{\begin{math}
	\left[\begin{array}{ll}#1\end{array}\right]\end{math}}
\newcommand{\Coref}[1]{\fbox{\tiny #1}}

\newcommand{\domlink}[1]{\begin{picture}(0,10)(0,0)
\multiput(0,0)(0,-2){5}{\line(0,-1){1}}
\put(-1,-5){\makebox(0,0)[r]{\treelabel{#1}}}
\end{picture}}

\newcommand{\rdomtree}[4]{\begin{picture}(30,22)(15,10)
\put(0,0){\line(3,2){15}}\put(7,5){\makebox(0,0)[br]{\treelabel{#1}}}
\put(15,10){\line(3,-2){15}}\put(23,5){\makebox(0,0)[bl]{\treelabel{#2}}}
\put(0,-1){#3}
\put(30,-9){#4}
\put(30,0){\domlink{D}}
\end{picture}}

\newcommand{\ldomtree}[4]{\begin{picture}(30,22)(15,10)
\put(0,0){\line(3,2){15}}\put(7,5){\makebox(0,0)[br]{\treelabel{#1}}}
\put(15,10){\line(3,-2){15}}\put(23,5){\makebox(0,0)[bl]{\treelabel{#2}}}
\put(0,-9){#3}
\put(30,-1){#4}
\put(0,0){\domlink{D}}
\end{picture}}

\newcommand{\idtree}[4]{\begin{picture}(30,10)(15,10)
\put(0,0){\line(3,2){15}}\put(7,5){\makebox(0,0)[br]{\treelabel{#1}}}
\put(15,10){\line(3,-2){15}}\put(23,5){\makebox(0,0)[bl]{\treelabel{#2}}}
\put(0,-0.5){#3}
\put(30,-0.5){#4}
\end{picture}}

\title{Compilation of HPSG to TAG\thanks{We would like to thank
	 A.~Abeill{\'e}, D.~Flickinger, A.~Joshi, T.~Kroch, O.~Rambow,
         I.~Sag and H.~Uszko\-reit
	 for valuable comments and discussions. The research underlying
	 the paper was supported by research grants from the German
	 Bundesministerium f\"ur Bildung, Wissenschaft, Forschung und
         Technologie (BMBF) to the DFKI projects {\sc Disco},
         FKZ~ITW~9002~0, {\sc Paradice},  FKZ~ITW~9403 and the {\sc
         VerbMobil} project, FKZ 01 IV 101 K/1, and by the Center for
         Cognitive Science at Ohio State University.}
}

\author{Robert~Kasper \\
        Dept. of Linguistics \\
        Ohio State University\\
        222 Oxley Hall\\
        Columbus, OH 43210 \\ U.S.A. \\
        kasper@ling.ohio-state.edu
        \And
        Bernd~Kiefer~~~Klaus~Netter\\
        Deutsches Forschungszentrum\\
        f\"ur K\"unstliche Intelligenz, GmbH\\
        Stuhlsatzenhausweg 3\\
        66123 Saarbr\"ucken \\ Germany \\
        \{kiefer$|$netter\}@dfki.uni-sb.de\\
        \And
	K.~Vijay-Shanker \\
        CIS Dept.\\
        University of Delaware\\
        Newark, DE 19716\\
        U.S.A \\
        vijay@cis.udel.edu}

\begin{document}

\maketitle

\begin{abstract}

We present an implemented compilation algorithm that translates
HPSG into lexicalized feature-based TAG, relating concepts of the two
theories. While HPSG has a more elaborated principle-based theory of
possible phrase structures, TAG provides the means to represent
lexicalized structures more explicitly.  Our objectives are met by
giving clear definitions that determine the projection of structures
from the lexicon, and identify ``maximal'' projections, auxiliary trees
and foot nodes.

\end{abstract}

\section{Introduction}

Head Driven Phrase Structure Grammar (HPSG) and Tree Adjoining Grammar
(TAG) are two frameworks which so far have been largely pursued in
parallel, taking little or no account of each other. In this paper we
will describe an algorithm which will compile HPSG grammars,
obeying certain constraints, into TAGs. However, we are not only
interested in mapping one formalism into another, but also in exploring
the relationship between concepts employed in the two frameworks.

HPSG is a feature-based grammatical framework which is characterized
by a modular specification of linguistic generalizations through
extensive use of principles and lexicalization of grammatical
information.  Traditional grammar rules are generalized to schemata
providing an abstract definition of grammatical relations, such as
head-of, complement-of, subject-of, adjunct-of, etc. Principles, such
as the Head-Feature-, Valence-, Non-Local- or Semantics-Principle,
determine the projection of information from the lexicon and
recursively define the flow of information in a global structure.
Through this modular design, grammatical descriptions are broken down
into minimal structural units referring to local trees of depth one,
jointly constraining the set of well-formed sentences.

In HPSG, based on the concept of ``head-domains'', {\em local} relations
(such as complement-of, adjunct-of) are defined as those that are
realized within the domain defined by the syntactic head.
This domain is usually the maximal projection of the head,
but it may be further extended in some cases, such as raising constructions.
In contrast,
filler-gap relations are considered {\em non-local}.
This local vs. non-local distinction in HPSG cuts across the relations
that are localized in TAG via the domains defined by elementary trees.
Each elementary tree typically represents all of the arguments
that are dependent on a lexical functor.
For example, the complement-of and filler-gap relations are localized in TAG,
whereas the adjunct-of relation is not.

Thus, there is a fundamental distinction between the different
notions of localization that have been assumed in the two frameworks.
If, at first sight, these frameworks seem to involve a radically
different organization of grammatical relations, it is natural to
question whether it is possible to compile one into the other in a
manner faithful to both, and more importantly, why this compilation is
being explored at all.  We believe that by combining the two
approaches both frameworks will profit.

{}From the HPSG perspective, this compilation offers the potential to
improve processing efficiency.
HPSG is a ``lexicalist'' framework, in the sense
that the lexicon contains the information that determines
which specific categories can be combined.
However, most HPSG grammars are not lexicalized in the stronger
sense defined by Schabes et.al.~\cite{saj88}, where lexicalization
means that each elementary structure in the
grammar is anchored by some lexical item.
For example, HPSG typically assumes a rule schema which combines
a subject phrase (e.g. NP) with a head phrase (e.g. VP),
neither of which is a lexical item.
Consider a sentence involving a
transitive verb which is derived by applying two rule schemata, reducing
first the object and then the subject.
In a standard HPSG derivation, once the head verb
has been retrieved, it must be computed that these two rules (and no
other rules) are applicable,
and then information about the complement and subject constituents is projected
from the lexicon according to the constraints on each rule schema.
On the other hand, in a lexicalized TAG derivation,
a tree structure corresponding to the combined instantiation of these
two rule schemata is directly retrieved along with the lexical
item for the verb.
Therefore, a procedure that compiles HPSG to TAG can be seen
as performing significant portions of an HPSG derivation at
compile-time, so that the structures projected from lexical items
do not need to be derived at run-time.
The compilation to TAG provides a
way of producing a strongly lexicalized grammar which
is equivalent to the original HPSG, and we expect this
lexicalization to yield a computational benefit in parsing
(cf.~\cite{Schabes-Joshi90}).

This compilation strategy also raises several issues of theoretical interest.
While TAG belongs to a class of mildly context-sensitive grammar formalisms
\cite{Joshi},
the generative capacity of the formalism underlying HPSG
(viz., recursive constraints over typed feature structures)
is unconstrained, allowing any recursively enumerable language
to be described.  In HPSG the constraints necessary
to characterize the class of natural languages are stated
within a very expressive formalism, rather than built into the
definition of a more restrictive formalism, such as TAG.
Given the greater
expressive power of the HPSG formalism, it will not be possible to
compile an aribitrary HPSG grammar into a TAG grammar.
However, our compilation algorithm shows that particular HPSG
grammars may contain constraints which have the effect
of limiting the generative capacity to that of
a mildly context-sensitive language.\footnote{We are only considering
a syntactic fragment of HPSG here.  It is not clear whether the semantic
components of HPSG can also be compiled into a more constrained formalism.
}
Additionally, our work provides a new perspective
on the different types of constituent combination in HPSG,
enabling a classification of schemata and principles in terms
of more abstract functor-argument relations.

\comment{Current work in the TAG framework does not offer modular and explicit
means for constraining the internal structure of elementary trees
according to principled generalizations.  In deriving elementary trees
from a HPSG source specification, we provide a principled way of
determining the content of the elementary trees (e.g., what to project
from lexical items and when to stop the projection).}

{}From a TAG perspective, using concepts employed in the HPSG framework,
we provide an explicit method of determining the content of the
elementary trees (e.g., what to project from lexical items and when to
stop the projection) from an HPSG source specification. This also
provides a method for deriving the distinctions between initial and
auxiliary trees, including the identification of foot nodes in auxiliary
trees.  Our answers, while consistent with basic tenets of traditional TAG
analyses, are general enough to allow an alternate linguistic theory,
such as HPSG, to be used as a basis for deriving a TAG. In this manner,
our work also serves to investigate the utility of the TAG framework
itself as a means of expressing different linguistic theories and
intuitions.

In the following we will first briefly describe the basic constraints we
assume for the HPSG input grammar and the resulting form of TAG.  Next
we describe the essential algorithm that determines the projection of
trees from the lexicon, and give formal definitions of
auxiliary tree and  foot node.  We then show how the
computation of ``sub-maximal'' projections can be triggered and carried
out in a two-phase compilation.

\section{Background}
\label{tag}

As the target of our translation we assume a Lexicalized Tree-Adjoining
Grammar~(LTAG), in which every elementary tree is anchored by a lexical
item~\cite{saj88}.

We do not assume atomic labelling of nodes, unlike traditional TAG,
where the root and foot nodes of an auxiliary tree are assumed to be
labelled identically.  Such trees are said to factor out recursion.
However, this identity itself isn't sufficient to identify foot nodes,
as more than one frontier node may be labelled the same as the root.
Without such atomic labels in HPSG, we are forced to address this
issue, and present a solution that is still consistent with the notion
of factoring recursion.

Our translation process yields a lexicalized feature-based TAG
\cite{vj88} in which feature structures are associated with nodes in the
frontier of trees and two feature structures (top and bottom) with nodes
in the interior.  Following \cite{v92}, the relationships between such
top and bottom feature structures represent underspecified domination
links.  Two nodes standing in this domination relation could become the
same, but they are necessarily distinct if adjoining takes place.
Adjoining separates them by introducing the path from the root to the
foot node of an auxiliary tree as a further specification of the
underspecified domination link.

For illustration of our compilation, we consider an extended HPSG
following the specifications in~\cite{PS2}[404ff].  The rule schemata
include rules for complementation (including head-subject and
head-complement relations), head-adjunct, and filler-head relations.

The following rule schemata cover the combination of heads with subjects
and other complements respectively as well as the adjunct
constructions.\footnote{We abstract from quite a number of properites
and use the following abbreviations for feature names:
\attr{S}=\attr{SYNSEM}, \attr{L}=\attr{LOCAL}, \attr{C}=\attr{CAT},
\attr{N-L}=\attr{NON-LOCAL}, \attr{D}=\attr{DTRS}.}

\medskip
{\it Head-Subj-Schema}

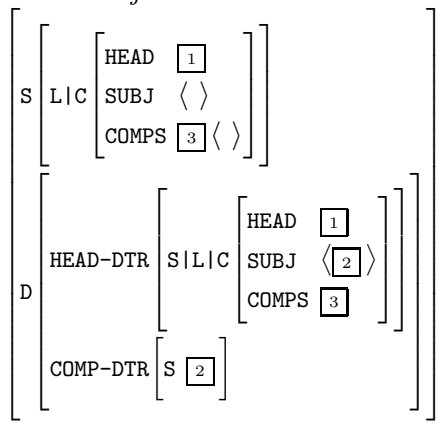
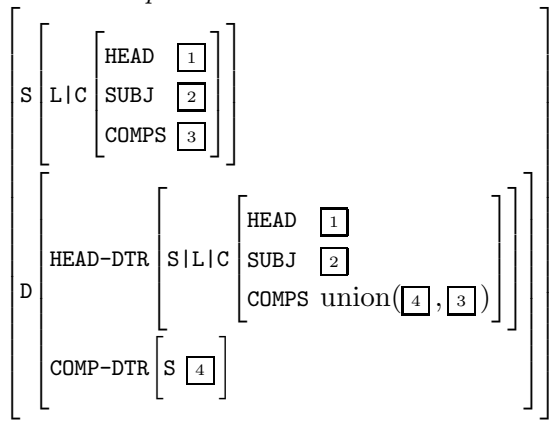
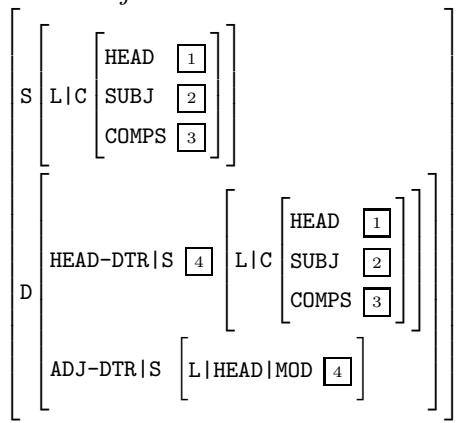
\avm{\avl{S}{\avm
                {\avl{L|C}{\avm
                        {\avl{HEAD}{\tind{1}}
                         \avl{SUBJ}{\eyl}
                         \avl{COMPS}{\tind{3}\eyl}}}}}
     \avl{D}{\avm
                {\avl{HEAD-DTR}{\avm
                	{\avl{S|L|C}{\avm
                                        {\avl{HEAD}{\tind{1}}
                                         \avl{SUBJ}{\nel{\tind{2}}}
                                         \avl{COMPS}{\tind{3}}}}}}
                \avl{COMP-DTR}{\avm{\avl{S}{\tind{2}}}}}}}

\medskip

{\it Head-Comps-Schema}

\avm{\avl{S}{\avm
                {\avl{L|C}{\avm
                        {\avl{HEAD}{\tind{1}}
                         \avl{SUBJ}{\tind{2}}
                         \avl{COMPS}{\tind{3}}}}}}
     \avl{D}{\avm
                {\avl{HEAD-DTR}{\avm
                	{\avl{S|L|C}{\avm
                                        {\avl{HEAD}{\tind{1}}
                                         \avl{SUBJ}{\tind{2}}
                                         \avl{COMPS}{{\rm
union}({\tind{4},\tind{3})}}}}}}
                \avl{COMP-DTR}{\avm
                	{\avl{S}{\tind{4}}}}}}}

\medskip

{\it Head-Adjunct-Schema}

\avm{\avl{S}{\avm
                {\avl{L|C}{\avm
                        {\avl{HEAD}{\tind{1}}
                         \avl{SUBJ}{\tind{2}}
                         \avl{COMPS}{\tind{3}}}}}}
     \avl{D}{\avm
                {\avl{HEAD-DTR|S}{\tind{4}~\avm
                        {\avl{L|C}{\avm
                                {\avl{HEAD}{\tind{1}}
                                 \avl{SUBJ}{\tind{2}}
                                 \avl{COMPS}{\tind{3} }}}}}
                \avl{ADJ-DTR|S}{\avm
                        {\avl{L|HEAD|MOD}{\tind{4}}}}}}}

\medskip

We assume a slightly modified and constrained treatment of non-local
dependencies (\attr{SLASH}), in which empty nodes are eliminated and
a lexical rule is used instead. While \attr{SLASH} introduction is based on
the standard filler-head schema, \attr{SLASH} percolation is essentially
constrained to the \attr{HEAD} spine.

\medskip

{\it Head-Filler-Schema}

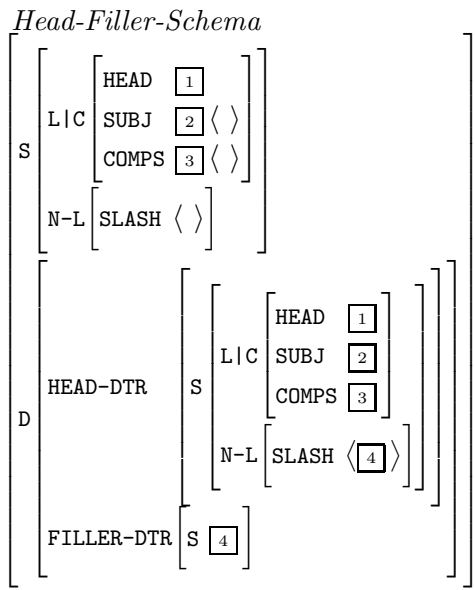
\avm{\avl{S}{\avm
                {\avl{L|C}{\avm
                        	{\avl{HEAD}{\tind{1}}
				 \avl{SUBJ}{\tind{2}\eyl}
				 \avl{COMPS}{\tind{3}\eyl}}}
		 \avl{N-L}{\avm
                        {\avl{SLASH}{\eyl}}}}}
     \avl{D}{\avm
                {\avl{HEAD-DTR}{\avm
                	{\avl{S}{\avm
                        	{\avl{L|C}{\avm
                                        {\avl{HEAD}{\tind{1}}
                                         \avl{SUBJ}{\tind{2}}
                                         \avl{COMPS}{\tind{3}}}}
				 \avl{N-L}{\avm
			 		{\avl{SLASH}{\nel{\tind{4}}}}}}}}}
                \avl{FILLER-DTR}{\avm
                	{\avl{S}{\tind{4}}}}}}}

\medskip

\attr{SLASH} termination is accounted for by a lexical rule, which
removes an element from one of the valence lists (\attr{COMPS} or
\attr{SUBJ}) and adds it to the \attr{SLASH} list.

\medskip

{\samepage
{\it Lexical~Slash-Termination-Rule}

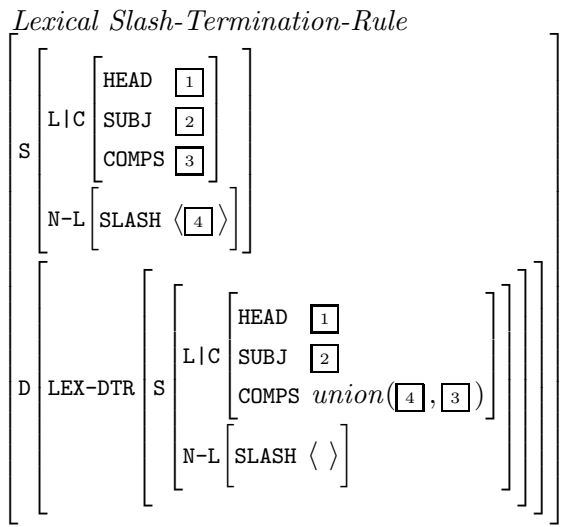
\nopagebreak
\avm{\avl{S}{\avm
                {\avl{L|C}{\avm
                        	{\avl{HEAD}{\tind{1}}
				 \avl{SUBJ}{\tind{2}}
				 \avl{COMPS}{\tind{3}}}}
		 \avl{N-L}{\avm
                        {\avl{SLASH}{\nel{\tind{4}}}}}}}
     \avl{D}{\avm
                {\avl{LEX-DTR}{\avm
                	{\avl{S}{\avm
                        	{\avl{L|C}{\avm
                                        {\avl{HEAD}{\tind{1}}
                                         \avl{SUBJ}{\tind{2}}
                                         \avl{COMPS}{union(\tind4,\tind{3})}}}
				\avl{N-L}{\avm
			 		{\avl{SLASH}{\eyl}}}}}}}}}}

}
\medskip

The percolation of \attr{SLASH} across head domains is lexically
determined.  Most lexical items will be specified as having an
empty \attr{SLASH} list. Bridge verbs (e.g., equi verbs such as {\em want})
or other heads allowing extraction
out of a complement share their own \attr{SLASH} value with the
\attr{SLASH} of the respective complement.\footnote{
We choose such a lexicalized approach, because it will allow us to
maintain a restriction that every TAG tree resulting from the
compilation must be rooted in a non-emtpy lexical item. The approach
will account for extraction of complements out of complements, i.e.,
along paths corresponding to chains of government relations.

As far as we can see, the only limitation arising from the percolation
of \attr{SLASH} only along head-projections is on extraction out of
adjuncts, which may be desirable for some languages like English.
On the other hand,
these constructions would have to be treated by multi-component TAGs,
which are not covered by the intended interpretation of the compilation
algorithm anyway.}

\medskip

{\it Equi and Bridge~Verb}

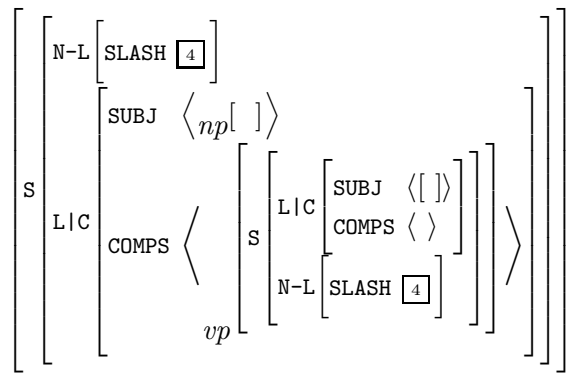
\avm{\avl{S}{\avm
                {\avl{N-L}{\avm
                        {\avl{SLASH}{\tind{4}}}}
		 \avl{L|C}{\avm{
			\avl{SUBJ}{\nel{\etavm{np}}}
                        \avl{COMPS}{\nel{
			 \tavm{vp}{\avl{S}
		         {\avm{\avl{L|C}{\avm{
				\avl{SUBJ}{\nel{\eavm{}}}
				\avl{COMPS}{\eyl}}}
			       \avl{N-L}{\avm{\avl{SLASH}{\tind{4}}
                                             }}}}}}}}}}}}

\medskip

Finally, we assume that rule schemata and principles have been compiled
together (automatically or manually) to yield more specific subtypes
of the schemata. This does not involve a loss of generalization but simply
means a further refinement of the type hierarchy. LP constraints could be
compiled out beforehand or during the compilation of TAG structures, since the
algorithm is lexicon driven.

\section{Algorithm}

\subsection{Basic Idea}

While in TAG all arguments related to a particular functor are
represented in one elementary tree structure, the `functional
application' in HPSG is distributed over the phrasal schemata, each of
which can be viewed as a partial description of a local tree.
Therefore we have to identify which constituents in a phrasal schema
count as functors and arguments. In TAG different functor argument
relations, such as head-complement, head-modifier etc., are
represented in the same format as branches of a trunk projected from a
lexical anchor. As mentioned, this anchor is not always equivalent to
the HPSG notion of a head; in a tree projected from a modifier, for
example, a non-head (\attr{ADJUNCT-DTR}) counts as a functor.  We
therefore have to generalize over different types of daughters in HPSG
and define a general notion of a functor. We compute the
functor-argument structure on the basis of a general {\em selection\/}
relation.
Following \cite{Kasper}\footnote{The algorithm presented here extends
and refines the approach described by~\cite{Kasper} by stating more
precise criteria for the projection of features, for the termination
of the algorithm, and for the determination of those structures which
should actually be used as elementary trees.
},
we adopt the notion of a {\em
selector daughter\/} (SD), which contains a {\em selector feature\/}
(SF) whose value constrains the argument (or non-selector) daughter
(non-SD).\footnote{Note that there might be mutual selection (as in
the case of the specifier-head-relations proposed in~\cite{PS2}[44ff]).
If there is mutual selection, we have to stipulate one of the
daughters as the SD.  The choice made would not effect the correctness
of the compilation.} For example, in a head-complement structure, the
SD is the \attr{HEAD-DTR}, as it contains the list-valued feature
\attr{COMPS} (the SF) each of whose elements selects a
\attr{COMP-DTR}, i.e., an element of the \attr{COMPS} list is
identified with the \attr{SYNSEM} value of a \attr{COMP-DTR}.

We assume that a {\em reduction} takes place along with selection.
Informally, this means that if \attr{F} is the selector feature for some
schema, then the value (or the element(s) in the list-value) of \attr{F}
that selects the non-SD(s) is not contained in the \attr{F} value of the
mother node.  In case \attr{F} is list-valued, we assume that the rest
of the elements in the list (those that did not select any daughter) are
also contained in the \attr{F} at the mother node.  Thus we say that
\attr{F} has been reduced by the schema in question.

The compilation algorithm assumes that all HPSG schemata will satisfy
the condition of simultaneous selection and reduction, and that each
schema reduces at least one \SF.  For the head-complement- and
head-subject-schema, these conditions follow from the Valence
Principle, and the SFs are \attr{COMPS}
and \attr{SUBJ}, respectively.  For the head-adjunct-schema, the
\attr{ADJUNCT-DTR} is the SD, because it selects the \attr{HEAD-DTR}
by its \attr{MOD} feature.  The \attr{MOD} feature is reduced, because
it is a head feature, whose value is inherited only from the
\attr{HEAD-DTR} and not from the \attr{ADJUNCT-DTR}.  Finally, for the
filler-head-schema, the \attr{HEAD-DTR} is the SD, as it selects the
\attr{FILLER-DTR} by its \attr{SLASH} value, which is bound off, not
inherited by the mother, and therefore reduced.

We now give a general description of the compilation process.
Essentially, we begin with a lexical description and project phrases
by using the schemata to reduce the selection information specified by the
lexical type.

\begin{description}

\item[Basic Algorithm] Take a lexical type L and initialize by creating
      a node with this type. Add a node $n$ dominating this node.

      For any schema S in which specified \SFs\ of $n$ are reduced, try
      to instantiate S with $n$ corresponding to the \SD\ of S. Add another
      node $m$ dominating the root node of the instantiated schema.
      (The domination links are
      introduced to allow for the possibility of adjoining.)  Repeat
      this step (each time with $n$ as the root node of the tree)
      until no further reduction is possible.

\end{description}

We will fill in the details below in the following order: what
information to raise across domination links (where adjoining may take
place), how to determine auxiliary trees (and foot nodes),
and when to terminate the projection.

We note that the trees produced have a {\em trunk} leading from the
lexical anchor (node for the given lexical type) to the root. The nodes
that are siblings of nodes on the trunk, the selected daughters, are not
elaborated further and serve either as foot nodes or substitution
nodes.

\subsection{Raising Features Across Domination Links}\label{raising}

Quite obviously, we {\em must\/} raise the SFs across domination links,
since they determine
the applicability of a schema and licence the instantiation of an \SD. If
no \SF\ were raised, we would lose all information about
the saturation status of a functor, and the algorithm would terminate
after the first iteration.

There is a danger in raising more than the \SFs. For example, the
head-subject-schema in German would typically constrain a verbal head to
be finite. Raising \attr{HEAD} features would block its application to
non-finite verbs and we would not produce the trees required for
raising-verb adjunction. This is again because {\em
heads\/} in HPSG are not equivalent to {\em lexical anchors\/} in TAG,
and that other local properties of the top and bottom of a domination link
could differ.  Therefore \attr{HEAD} features
and other \attr{LOCAL} features cannot, in general, be raised across
domination links, and we assume for now that only the SFs are raised.

Raising all SFs produces only fully saturated elementary trees and
would require the root and foot of any auxiliary tree
to share all SFs, in order to be compatible with the SF values
across any domination links where adjoining can take place.
This is too strong a condition and will not allow the
resulting TAG to generate all the trees derivable with the given HPSG
(e.g., it would not allow unsaturated VP complements).
In \S~\ref{xxx} we address this concern by using a
multi-phase compilation. In the first phase, we raise all the \SFs.

\subsection{Detecting Auxiliary Trees and Foot Nodes}

Traditionally, in TAG, auxiliary trees are said to be minimal recursive
structures that have a foot node (at the frontier) labelled identical to
the root.  As such category labels ($S, NP$ etc.) determine where an
auxiliary tree can be adjoined, we can informally think of these labels
as providing selection information corresponding to the SFs of HPSG.
Factoring of recursion can then be viewed as saying that auxiliary trees
define a path (called the {\em spine}) from the root to the foot where the
nodes at extremities have the same selection information. However, a
closer look at TAG shows that this is an oversimplification. If we take
into account the adjoining constraints (or the top and bottom feature
structures), then it appears that the root and foot share only some
selection information.

Although the encoding of selection information by SFs in HPSG is
somewhat different than that traditionally employed in TAG, we also
adopt the notion that the extremities of the spine in an auxiliary tree
share some part (but not necessarily all) of the selection information.
Thus, once we have produced a tree, we examine the root and the nodes in
its frontier. A tree is an auxiliary tree if the root and some
frontier node (which becomes the foot node) have some non-empty
SF value in common. Initial trees are those that have no such
frontier nodes.

\medskip

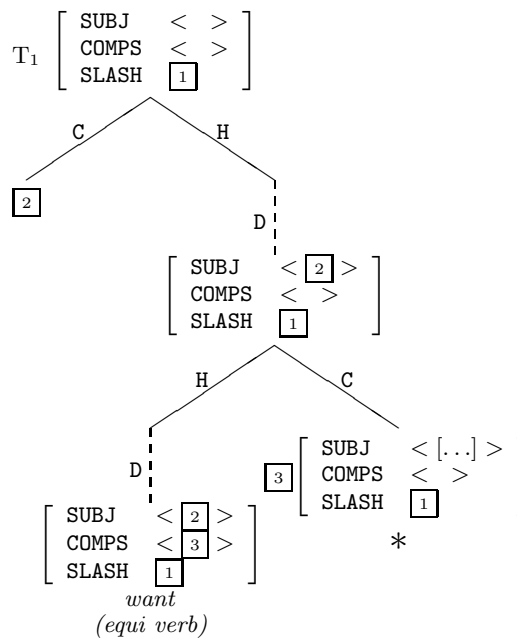
\begin{center}

{\small

\begin{picture}(58,78)(0,12)
\put(0,85){\makebox(0,0){$\mbox{T}_1$}}
\put(15,85){\makebox(0,0)[b]{
\AVM{
{\tt SUBJ}&\Sequence{~}\\
{\tt COMPS}&\Sequence{~}\\
{\tt SLASH}&\fbox{\tiny 1}
}}}
\put(15,80){\rdomtree{C}{H}{\makebox(0,0)[t]{\fbox{\tiny 2}}}
{\begin{picture}(100,100)(100,100)
\put(100,100){\makebox(0,0)[t]{\AVM{
{\tt SUBJ}&\Sequence{\fbox{\tiny 2}}\\
{\tt COMPS}&\Sequence{~}\\
{\tt SLASH}&\fbox{\tiny 1}
}}}
\put(100,89){\ldomtree{H}{C}{\makebox(0,0)[t]{\begin{tabular}{c}
\AVM{
{\tt SUBJ}&\Sequence{\fbox{\tiny 2}}\\
{\tt COMPS}&\Sequence{\fbox{\tiny 3}}\\
{\tt SLASH}&\fbox{\tiny 1}
}\\[0.4ex]
{\em want\/}\\
{\em (equi verb)}
\end{tabular}}}
{\makebox(0,0)[t]{\begin{tabular}{c}\fbox{\tiny 3}\AVM{
{\tt SUBJ}&\Sequence{[\ldots]}\\
{\tt COMPS}&\Sequence{~}\\
{\tt SLASH}&\fbox{\tiny 1}
}\\[4ex]{\Large $\ast$}\end{tabular}}}}
\end{picture}}}
\end{picture}
}
\end{center}

In the trees shown, nodes detected as foot nodes are marked with $\ast$.
Because of the \attr{SUBJ} and \attr{SLASH} values, the \attr{HEAD-DTR}
is the foot of T$_2$ below (anchored by an adverb) and \attr{COMP-DTR}
is the foot of T$_3$ (anchored by a raising verb). Note that in the tree
T$_1$ anchored by an equi-verb, the foot node is detected because the
\attr{SLASH} value is shared, although the \attr{SUBJ} is not. As
mentioned, we assume that bridge verbs, i.e., verbs which allow
extraction out of their complements, share their \attr{SLASH} value with
their clausal complement.

\subsection{Termination}

Returning to the basic algorithm, we will now consider the issue of
termination, i.e., how much do we need to reduce as we project a tree
from a lexical item.

Normally, we expect a SF with a specified value to be reduced fully
to an empty list by a
series of applications of rule schemata.  However, note that the
\attr{SLASH} value is unspecified at the root of the trees T$_2$ and
T$_3$.  Of course, such nodes would still unify with the SD of the
filler-head-schema (which reduces \attr{SLASH}), but applying this
schema could lead to an infinite recursion.  Applying a reduction to an
unspecified SF is also linguistically unmotivated as it would imply that
a functor could be applied to an argument that it never explicitly
selected.

However, simply blocking the reduction of a SF whenever its value is
unspecified isn't sufficient. For example, the root of T$_2$ specifies
the \attr{SUBJ} to be a non-empty list. Intuitively, it would not be
appropriate to reduce it further, because the lexical anchor (adverb)
doesn't semantically license the \attr{SUBJ} argument itself.  It merely
constrains the modified head to have an unsaturated \attr{SUBJ}.

\medskip

{\small

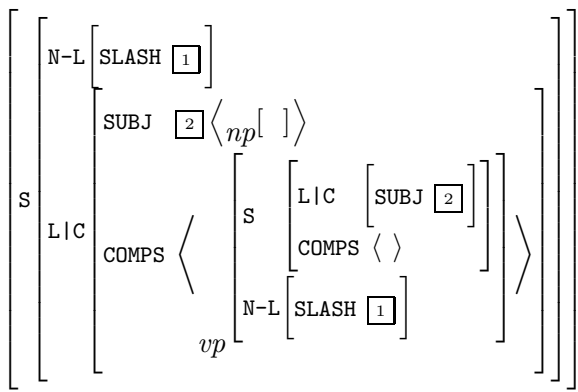
\begin{picture}(55,46)(51,45)
\put(65,85){\makebox(0,0){$\mbox{T}_2$}}
\put(80,85){\makebox(0,0)[b]{\AVM{
{\tt SUBJ}&{\fbox{\tiny 2}}\\
{\tt COMPS}&{\Sequence{~}}\\
{\tt SLASH}&{\fbox{\tiny 3}}
}}}
\put(65,70){\line(3,2){15}}\put(71.5,75){\makebox(0,0)[br]{\tt A}}
\put(80,80){\line(3,-2){15}}\put(88.5,75){\makebox(0,0)[bl]{\tt H}}
\put(95,69){\makebox(0,0)[t]{\begin{tabular}{c}\Coref{1}
\AVM{
{\tt SUBJ}&{\fbox{\tiny 2}\Sequence{\ \avm{~}}}\\
{\tt COMPS}&{\Sequence{~}}\\
{\tt SLASH}&{\fbox{\tiny 3}}
}\\[4ex]{\Large $\ast$}
\end{tabular}}}
\multiput(65,70)(0,-2){6}{\line(0,-1){1}} 
\put(66,64){\makebox(0,0)[l]{\tt D}}
\put(65,59){\makebox(0,0)[t]{\begin{tabular}{c}\AVM{
{\tt SUBJ}&\Sequence{~}\\
{\tt COMPS}&\Sequence{~}\\
{\tt SLASH}&\Sequence{~}\\
{\tt MOD}&{\fbox{\tiny 1}}\\}\\[0.4ex]
{\em VP-adverb}
\end{tabular}}}
\end{picture}
}

\bigskip

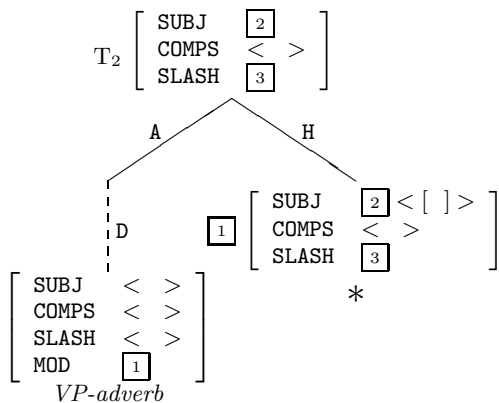
{\em Raising~Verb (and Infinitive Marker {\em to\/})}

\avm{\avl{S}{\avm
                {\avl{N-L}{\avm
                        {\avl{SLASH}{\tind{1}}}}
		 \avl{L|C}{\avm{\avl{SUBJ}{\tind{2}{\nel{\etavm{np}}}}
                      	\avl{COMPS}{\nel{
			 \tavm{vp}{
			   \avl{S}{\avm{\avl{L|C}{\avm{\avl{SUBJ}{\tind{2}}}}
				        \avl{COMPS}{\eyl}}}
			   \avl{N-L}{\avm{\avl{SLASH}{\tind{1}}
                                              }}}}}}}}}}

\medskip

{\small

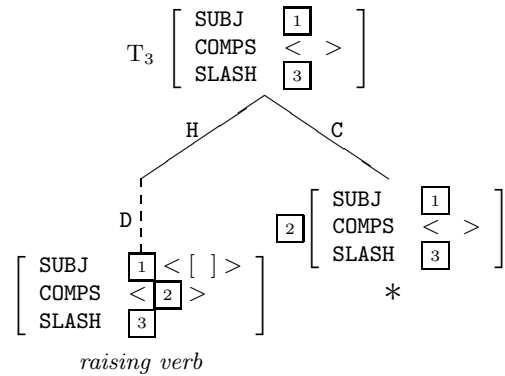
\begin{picture}(56,44)(-18,47)
\put(0,85){\makebox(0,0){$\mbox{T}_3$}}
\put(15,85){\makebox(0,0)[b]{
\AVM{
{\tt SUBJ}&{\Coref{1}}\\
{\tt COMPS}&\Sequence{~}\\
{\tt SLASH}&\Coref{3}
}}}
\put(15,80){\ldomtree{H}{C}{\makebox(0,0)[t]{\begin{tabular}{c}
\AVM{
{\tt SUBJ}&{\Coref{1}\Sequence{\ \avm{~}}}\\
{\tt COMPS}&\Sequence{\Coref{2}}\\
{\tt SLASH}&\Coref{3}
}\\[4ex]
{\em raising verb}
\end{tabular}}
}
{
\makebox(0,0)[t]{\begin{tabular}{c}
\Coref{2}\AVM{
{\tt SUBJ}&{\Coref{1}}\\
{\tt COMPS}&\Sequence{~}\\
{\tt SLASH}&\Coref{3}
}\\[4ex]{\Large $\ast$}
\end{tabular}
}}}
\end{picture}

}

\medskip

To motivate our termination criterion, consider the adverb tree and
the asterisked node (whose \attr{SLASH} value is shared with
\attr{SLASH} at the root).
Being a non-trunk node, it will either be a foot or a substitution node.
In either case, it will eventually be unified with some node in another
tree.
If that other node has a reducible \attr{SLASH} value, then we know
that the reduction takes place in the other tree, because the
\attr{SLASH} value must have been raised across the domination link
where adjoining takes place.  As the same \attr{SLASH} (and likewise
\attr{SUBJ}) value should not be reduced in both trees, we state our
termination criteria as follows:

\begin{description}
\item[Termination Criterion] The value of an SF \attr{F} at the root node
of a tree is not reduced further if it is an empty list,
or if it is shared with the value of \attr{F}
at some non-trunk node in the frontier.
\end{description}

Note that because of this termination criterion, the adverb tree
projection will stop at this point. As the root shares some selector
feature values (\attr{SLASH} and \attr{SUBJ}) with a frontier node, this
node becomes the foot node. As observed above, adjoining this tree will
preserve these values across any domination links where it might be
adjoined; and if the values stated there are reducible then they will be
reduced in the other tree.  While auxiliary trees allow arguments
selected at the root to be realized elsewhere, it is never the case for
initial trees that an argument selected at the root can be realized
elsewhere, because by our definition of initial trees the selection of
arguments is not passed on to a node in the frontier.

We also obtain from this criterion a notion of {\em local
completeness\/}. A tree is locally complete as soon as all arguments
which it licenses and which are not licensed elsewhere are realized.
{\em Global completeness\/} is guaranteed because the notion of
``elsewhere'' is only and always defined for auxiliary trees, which have
to adjoin into an initial  tree.

\subsection{Additional Phases}\label{xxx}

Above, we noted that the preservation of some SFs
along a path (realized as a path from the root to the foot of an
auxiliary tree) does not imply that all SFs need to be
preserved along that path. Tree T$_1$ provides such an example, where a
lexical item, an equi-verb, triggers the reduction of an \SF\ by taking
a complement that is unsaturated for \attr{SUBJ} but never shares this
value with one of its own \SF\ values.

To allow for adjoining of auxiliary trees whose root and foot differ
in their SFs, we could produce a number of different trees
representing partial projections from each lexical anchor.  Each
partial projection could be produced by raising some subset of SFs
across each domination link, instead of raising all SFs.  However,
instead of systematically raising all possible subsets of SFs across
domination links, we can avoid producing a vast number of these
partial projections by using auxiliary trees to provide guidance in
determining when we need to raise only a particular subset of the SFs.

Consider T$_1$ whose root and foot differ in their SFs. From this we
can infer that a \attr{SUBJ} SF should not always be raised across
domination links in the trees compiled from this grammar.  However,
it is only useful to produce a tree in which
the \attr{SUBJ} value is not raised when the bottom of a
domination link has both a one element list as value for \attr{SUBJ}
and an empty \attr{COMPS} list. Having an empty \attr{SUBJ} list at
the top of the domination link would then allow for adjunction by
trees such as T$_1$.

This leads to the following multi-phase compilation algorithm. In the
first phase, all SFs are raised.  It is determined which trees are
auxiliary trees, and then the relationships between the SFs associated
with the root and foot in these auxiliary trees are recorded.  The
second phase begins with lexical types and considers the application
of sequences of rule schemata as before.  However, immediately after
applying a rule schema, the features at the bottom of a domination
link are compared with the foot nodes of auxiliary trees that have
differing SFs at foot and root.  Whenever the features are compatible
with such a foot node, the SFs are raised according to the
relationship between the root and foot of the auxiliary tree in
question.  This process may need to be iterated based on any new
auxiliary trees produced in the last phase.

\subsection{Example Derivation}

In the following we provide a sample derivation for the sentence

{\em (I know) what Kim wants to give to Sandy.\/}

Most of the relevant HPSG rule schemata and lexical entries necessary to
derive this sentence were already given above. For the noun phrases {\em
what\/}, {\em Kim\/} and {\em Sandy\/}, and the preposition {\em to\/}
no special assumptions are made.  We therefore only add the entry for
the ditransitive verb {\em give\/}, which we take to subcategorize for a
subject and two object complements.

\medskip

{\samepage
{\it Ditransitive~Verb}

\nopagebreak
\avm{\avl{S}{\avm
                {\avl{N-L}{\avm
                        {\avl{SLASH}{\eyl}}}
		 \avl{L|C}{\avm{\avl{SUBJ}{\nel{\etavm{np}}}
                        	\avl{COMPS}{\nel{
				 	\etavm{np}
                                        \etavm{pp}}}}}}}}
}
\medskip

{}From this lexical entry, we can derive in the first phase a fully
saturated initial tree by applying first the lexical slash-termination
rule, and then the head-complement-, head-subject and filler-head-rule.
Substitution at the nodes on the frontier would yield the string {\em
what Kim gives to Sandy\/}.

{\small

\begin{picture}(74,123)(0,-30)
\put(0,85){\makebox(0,0){$\mbox{T}_4$}}
\put(15,85){\makebox(0,0)[b]{
\AVM{
{\tt SUBJ}&\Sequence{~}\\
{\tt COMPS}&\Sequence{~}\\
{\tt SLASH}&\Sequence{~}
}}}
\put(15,80){\rdomtree{F}{H}
{\makebox(0,0)[t]{
\begin{tabular}{c}
\Coref{1}\\ [0.4ex]
{\em NP} \\
$\downarrow$\\
{\em what}
\end{tabular}
}}
{\begin{picture}(100,100)(100,100)
\put(100,100){\makebox(0,0)[t]{\AVM{
{\tt SUBJ}&\Sequence{~}\\
{\tt COMPS}&\Sequence{~}\\
{\tt SLASH}&\Sequence{\Coref{1}}
}}}
\put(100,89){\rdomtree{C}{H}
{\makebox(0,0)[t]{
\begin{tabular}{c}
\Coref{1}\\ [0.4ex]
{\em NP} \\
$\downarrow$\\
{\em Kim}
\end{tabular}
}}
{\begin{picture}(100,100)(100,100)
\put(100,100){\makebox(0,0)[t]{\AVM{
{\tt SUBJ}&\Sequence{\Coref{2}}\\
{\tt COMPS}&\Sequence{~}\\
{\tt SLASH}&\Sequence{\Coref{1}}
}}}
\put(100,89){\ldomtree{H}{C}
{\begin{picture}(100,100)(100,100)
\put(100,100){\makebox(0,0)[t]{\AVM{
{\tt SUBJ}&\Sequence{\Coref{2}}\\
{\tt COMPS}&\Sequence{\Coref{3}}\\
{\tt SLASH}&\Sequence{\Coref{1}}\\
}}}
\put(100,89){\line(0,-1){8}}
\put(99,84){\makebox(0,0)[r]{\tt LD}}
\put(100,79){\makebox(0,0)[t]{\begin{tabular}{c}
\AVM{
{\tt SUBJ}&\Sequence{\Coref{2}}\\
{\tt COMPS}&\Sequence{\Coref{1}\/,\Coref{3}}\\
{\tt SLASH}&\Sequence{~}
}\\[0.4ex]
{\em gives}
\end{tabular}}}
\end{picture}}
{\makebox(0,0)[t]{
\begin{tabular}{c}
\Coref{3}\\ [0.4ex]
{\em PP} \\
$\downarrow$\\
{\em to Sandy}
\end{tabular}
}}}
\end{picture}}}
\end{picture}}}
\end{picture}
}

\medskip

The derivations for the trees for the matrix verb {\em want\/} and for
the infinitival marker {\em to\/} (equivalent to a raising verb) were
given above in the examples T$_1$ and T$_3$. Note that the
\attr{SUBJ} feature is only reduced in the former, but not in the latter
structure.

In the second phase we derive from the entry for {\em give\/} another
initial tree (T$_5$) into which the auxiliary tree T$_1$ for {\em
want\/} can be adjoined at the topmost domination link.  We also
produce a second tree with similar properties for the infinitive marker
{\em to\/} (T$_6$).

{\small

\begin{picture}(59,110)(5,-39)
\put(15,65){\makebox(0,0){$\mbox{T}_5$}}
\put(30,65){\makebox(0,0)[b]{
\AVM{
{\tt SUBJ}&\Sequence{~}\\
{\tt COMPS}&\Sequence{~}\\
{\tt SLASH}&\Sequence{~}
}}}
\put(30,60){\idtree{F}{H}
{\makebox(0,0)[t]{
\begin{tabular}{c}
\Coref{1}\\ [0.4ex]
{\em NP} \\
$\downarrow$\\
{\em what}
\end{tabular}
}}
{\begin{picture}(100,100)(100,100)
\put(100,100){\makebox(0,0)[t]{\AVM{
{\tt SUBJ}&\Sequence{~}\\
{\tt COMPS}&\Sequence{~}\\
{\tt SLASH}&\Sequence{\Coref{1}}
}}}
\multiput(100,89)(0,-2){5}{\line(0,-1){1}}
\put(99,84){\makebox(0,0)[r]{\tt D}}
\put(100,79){\makebox(0,0)[t]{\AVM{
{\tt SUBJ}&\Sequence{\Coref{2}}\\
{\tt COMPS}&\Sequence{~}\\
{\tt SLASH}&\Sequence{\Coref{1}}
}}}
\put(100,68){\ldomtree{C}{H}
{\begin{picture}(100,100)(100,100)
\put(100,100){\makebox(0,0)[t]{\AVM{
{\tt SUBJ}&\Sequence{\Coref{2}}\\
{\tt COMPS}&\Sequence{\Coref{3}}\\
{\tt SLASH}&\Sequence{\Coref{1}}\\
}}}
\put(100,89){\line(0,-1){8}}
\put(99,84){\makebox(0,0)[r]{\tt LD}}
\put(100,79){\makebox(0,0)[t]{\begin{tabular}{c}
\AVM{
{\tt SUBJ}&\Sequence{\Coref{2}}\\
{\tt COMPS}&\Sequence{\Coref{3}\/,\Coref{1}}\\
{\tt SLASH}&\Sequence{~}
}\\[0.4ex]
{\em give}
\end{tabular}}}
\end{picture}}
{\makebox(0,0)[t]{
\begin{tabular}{c}
\Coref{3}\\ [0.4ex]
{\em PP} \\
$\downarrow$\\
{\em to Sandy}
\end{tabular}
}}}
\end{picture}}}
\end{picture}

}

{\small

\begin{picture}(56,60)(-18,28)
\put(0,85){\makebox(0,0){$\mbox{T}_6$}}
\put(16,90){\makebox(0,0)[t]{\AVM{
{\tt SUBJ}&\Sequence{~}\\
{\tt COMPS}&\Sequence{~}\\
{\tt SLASH}&\Sequence{\Coref{1}}
}}
\begin{picture}(100,100)(100,95)
\put(100,85){\domlink{D}}
\put(100,70){\makebox(0,0)[b]{
\AVM{
{\tt SUBJ}&{\Coref{1}}\\
{\tt COMPS}&\Sequence{~}\\
{\tt SLASH}&\Coref{1}
}}
\begin{picture}(100,100)(100,100)
\put(100,96){\ldomtree{H}{C}{\makebox(0,0)[t]{\begin{tabular}{c}
\AVM{
{\tt SUBJ}&{\Coref{1}\Sequence{\ \avm{~}}}\\
{\tt COMPS}&\Sequence{\Coref{2}}\\
{\tt SLASH}&\Coref{3}
}\\[4ex]
{\em to}
\end{tabular}}
}
{
\makebox(0,0)[t]{\begin{tabular}{c}
\Coref{2}\AVM{
{\tt SUBJ}&{\Coref{1}}\\
{\tt COMPS}&\Sequence{~}\\
{\tt SLASH}&\Coref{3}
}\\[4ex]{\Large $\ast$}
\end{tabular}
}}}
\end{picture}}
\end{picture}}
\end{picture}

}

By first adjoining the tree T${_6}$ at the topmost
domination link of T${_5}$ we obtain a structure T${_7}$ corresponding
to the substring {\em what ... to give to Sandy\/}.  Adjunction involves
the identification of the foot node with the bottom of the
domination link and identification of the root with top of the
domination link.  Since the domination link at the root of the adjoined
tree mirrors the properties of the adjunction site in the initial tree,
the properties of the domination link are preserved.

{\small

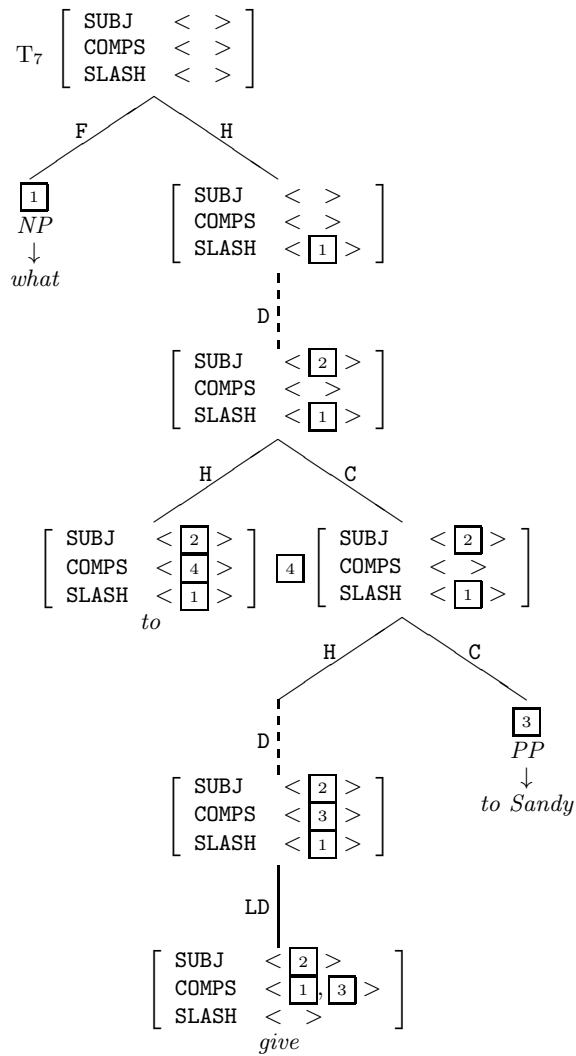
\begin{picture}(74,102)(0,-10)
\put(0,85){\makebox(0,0){$\mbox{T}_7$}}
\put(15,85){\makebox(0,0)[b]{
\AVM{
{\tt SUBJ}&\Sequence{~}\\
{\tt COMPS}&\Sequence{~}\\
{\tt SLASH}&\Sequence{~}
}}}
\put(15,80){\idtree{F}{H}
{\makebox(0,0)[t]{
\begin{tabular}{c}
\Coref{1}\\ [0.4ex]
{\em NP} \\
$\downarrow$\\
{\em what}
\end{tabular}}}
{\begin{picture}(100,100)(100,100)
\put(100,100){\makebox(0,0)[t]{\AVM{
{\tt SUBJ}&\Sequence{~}\\
{\tt COMPS}&\Sequence{~}\\
{\tt SLASH}&\Sequence{\Coref{1}}
}}}
\multiput(100,89)(0,-2){5}{\line(0,-1){1}}
\put(99,84){\makebox(0,0)[r]{\tt D}}
\put(100,80){\makebox(0,0)[t]{\AVM{
{\tt SUBJ}&\Sequence{\Coref{2}}\\
{\tt COMPS}&\Sequence{~}\\
{\tt SLASH}&\Sequence{\Coref{1}}
}}}
\put(100,69){\idtree{H}{C}
{\begin{picture}(100,100)(100,100)
\put(100,100){\makebox(0,0)[t]{\begin{tabular}{c}
\AVM{
{\tt SUBJ}&\Sequence{\Coref{2}}\\
{\tt COMPS}&\Sequence{\Coref{4}}\\
{\tt SLASH}&\Sequence{\Coref{1}}}
\\[0.4ex]
{\em to}
\end{tabular}
}}
\end{picture}}
{\begin{picture}(100,100)(100,100)
\put(100,100){\makebox(0,0)[t]{
\Coref{4}$\;$\AVM{
{\tt SUBJ}&\Sequence{\Coref{2}}\\
{\tt COMPS}&\Sequence{~}\\
{\tt SLASH}&\Sequence{\Coref{1}}
}}}
\put(100,89){\ldomtree{H}{C}
{\begin{picture}(100,100)(100,100)
\put(100,100){\makebox(0,0)[t]{\AVM{
{\tt SUBJ}&\Sequence{\Coref{2}}\\
{\tt COMPS}&\Sequence{\Coref{3}}\\
{\tt SLASH}&\Sequence{\Coref{1}}\\
}}}
\put(100,89){\line(0,-1){10}}
\put(99,84){\makebox(0,0)[r]{\tt LD}}
\put(100,79){\makebox(0,0)[t]{\begin{tabular}{c}
\AVM{
{\tt SUBJ}&\Sequence{\Coref{2}}\\
{\tt COMPS}&\Sequence{\Coref{1}\/,\Coref{3}}\\
{\tt SLASH}&\Sequence{~}
}\\[0.4ex]
{\em give}
\end{tabular}}}
\end{picture}}
{\makebox(0,0)[t]{
\begin{tabular}{c}
\Coref{3}\\ [0.4ex]
{\em PP} \\
$\downarrow$\\
{\em to Sandy}
\end{tabular}
}}}
\end{picture}}}
\end{picture}}}
\end{picture}

}

\vspace{28mm}

The final derivation step then involves the adjunction of the tree for
the equi verb into this tree, again at the topmost domination link.
This has the effect of inserting the substring {\em Kim wants} into
{\em what ... to give to Sandy\/}.

\section{Conclusion}

We have described how HPSG specifications can be compiled into TAG, in a
manner that is faithful to both frameworks.  This algorithm has been
implemented in Lisp and used to compile a significant fragment of a
German HPSG.  Work is in progress on compiling an English grammar
developed at CSLI.

This compilation strategy illustrates
how linguistic theories other than those previously explored within the
TAG formalism can be instantiated in TAG, allowing the association of
structures with an enlarged domain of locality with lexical items.
We have generalized the notion of factoring recursion in TAG,
by defining auxiliary trees in a way that is
not only adequate for our purposes,
but also provides a uniform treatment of extraction from
both clausal and non-clausal complements (e.g., VPs)
that is not possible in traditional TAG.

It should be noted that the results of our compilation will not always
conform to conventional linguistic assumptions often adopted in TAGs,
as exemplified by the auxiliary trees produced for equi verbs.
Also, as the algorithm does not currently include
any downward expansion from complement nodes on the frontier,
the resulting trees will sometimes be more
fractioned than if they had been specified directly in a TAG.

We are currently exploring the possiblity of compiling HPSG into an
extension of the TAG formalism, such as D-tree
grammars~\cite{Vijay95} or the UVG-DL formalism~\cite{Rambow}.
These somewhat more powerful formalisms appear to be adequate for some
phenomena, such as extraction out of adjuncts (recall~\S\ref{tag})
and certain kinds of scrambling, which our current method does not
handle.  More flexible methods of combining trees with dominance links
may also lead to a reduction in the number of trees that must be
produced in the second phase of our compilation.

There are also several techniques that we expect to lead to improved
parsing efficiency of the resulting TAG.  For instance, it is possible
to declare specific non-SFs which can be raised, thereby reducing the
number of useless trees produced during the multi-phase compilation.  We
have also developed a scheme to effectively organize the trees
associated with lexical items.

\vspace{-2mm}

\bibliographystyle{acl}

\setlength{\itemsep}{-2pt}

\end{document}